\documentclass[runningheads]{llncs}
\usepackage{graphicx}
\usepackage{cite}
\usepackage{amsmath,amssymb,amsfonts}
\usepackage{algorithmic}
\usepackage{graphicx}
\usepackage{textcomp}
\usepackage{xcolor}
\usepackage{picinpar}
\usepackage{multirow}
\usepackage{bm}
\usepackage{marvosym}

\begin{document}
\title{Vim-F: Visual State Space Model Benefiting from Learning in the Frequency Domain}
\titlerunning{Vim-F}
\author{Juntao Zhang\inst{1},Shaogeng Liu\inst{1}\thanks{Equal contribution.},Jun Zhou\inst{1},
Kun Bian\inst{2},You Zhou\inst{2},Jianning Liu\inst{1},Pei Zhang\inst{3},Bingyan Liu\inst{1}\thanks{ Corresponding author.},}
%
\authorrunning{J. Zhang et al.}
%

\institute{AMS, Beijing, China \\ \email{m201773066@alumni.hust.edu.cn}\\ \and
	School of Electronic Engineering\\ 
	Xidian University, Xi’an, China \and Coolanyp L.L.C.}
\maketitle              
\begin{abstract}
In recent years, State Space Models (SSMs) with efficient hardware-aware designs, known as Mamba deep learning models, have made significant progress in modeling long sequences. Compared to Convolutional Neural Networks (CNNs) and Vision Transformers (ViTs), Vision Mamba (ViM) methods have not yet achieved fully competitive performance. To enable SSMs to process image data, ViMs typically flatten 2D images into 1D sequences, inevitably ignoring some 2D local dependencies, thereby weakening the model's ability to interpret spatial relationships from a global perspective. We believe that the introduction of frequency domain information can enable ViM to achieve a better global receptive field during the scanning process. We propose a novel model called Vim-F, which employs pure Mamba encoders and scans in both the frequency and spatial domains. Moreover, considering that Mamba remains essentially a Recurrent Neural Network (RNN), we question the necessity of position embedding in ViM and remove it accordingly in Vim-F. Vim-F has good scalability. As far as we know, its variant Vim-F(CF) is the first ViM model to use a convolution-free ViM encoder. Another variant, Vim-F(H), introduces a linear attention mechanism. This reduces the model's sensitivity to the input sequence and achieves better performance.
\end{abstract}
\section{Introduction}
Visual representation learning stands as a pivotal research topic within the realm of computer vision. In the early era of deep learning, Convolutional Neural Networks (CNNs) occupied a dominant position. CNNs have several built-in inductive biases that make them well-suited to a wide variety of computer vision applications. In addition, CNNs are inherently efficient because when used in a sliding-window manner, the computations are shared. Inspired by the significant success of the Transformer architecture in the field of Natural Language Processing (NLP), researchers have recently applied Transformer to computer vision tasks. Except for the initial ``patch embedding" module, which splits an image into a sequence of patches, Vision Transformers (ViTs) do not introduce image-specific inductive biases and follow the original NLP Transformers design concept as much as possible. Compared to CNNs, ViTs typically exhibit superior performance, which can be attributed to the global receptive field and dynamic weights facilitated by the attention mechanism. However, the attention mechanism requires quadratic complexity in terms of image sizes, resulting in significant computational cost when addressing downstream dense prediction tasks, such as object detection and instance segmentation.\par
Derived from the classic Kalman filter model, modern State Space Models (SSMs) excel at capturing long-range dependencies and reap the benefits of parallel training. The Visual Mamba Models (ViMs) \cite{ref1,ref2,ref3,ref4}, which are inspired by recently proposed SSMs \cite{ref5,ref6}, utilize the Selective Scan Space State Sequential Model (S6) to compress previously scanned information into hidden states, effectively reducing quadratic complexity to linear. Most of these studies integrate the original SSM structure of Mamba into their basic models, thereby necessitating the conversion of 2D images into 1D sequences for SSM-based processing. However, due to the non-causal nature of visual data, flattening spatial data into one-dimensional tokens destroys local two-dimensional dependencies, impairing the model's capacity to accurately interpret spatial relationships. Vim \cite{ref1} addresses this issue by scanning in bidirectional horizontal directions, while VMamba \cite{ref2} adds vertical scanning, enabling each element in the feature map to integrate information from other locations in different directions. Subsequent works, such as LocalMamba \cite{ref3} and EfficientVMamba \cite{ref4}, have designed a series of novel scanning strategies in the spatial domain, aimed at capturing local dependencies while maintaining a global view. We believe that these strategies do not significantly enhance the model's ability to obtain a better global receptive field. These efforts have prompted us to focus on the value of scanning in the frequency domain.\par 
We know that the use of 2D Discrete Fourier transform (DFT) to convert feature maps into spectrograms does not alter their shapes. The value $P(x,y)$ at any point $(x,y)$ in the spectrogram depends on the entire original feature map. Therefore, scanning in the frequency domain ensures that the model always has a good global receptive field. On the other hand, without considering spectral shifts, the high-frequency components are located in the corners of the spectrogram, whereas the low-frequency components are centered. Scanning across the spectrogram typically alternates between accessing low and high frequencies, which may facilitate balanced modeling. Therefore, we adopt the most direct solution, which is to add the amplitude spectrum of the feature map to the original feature map, allowing the visual Mamba encoder to learn the fused semantic information. Using the Fast Fourier Transform (FFT) algorithm to calculate the amplitude spectrum not only introduces no trainable parameters but also has minimal impact on performance. To clearly demonstrate the effectiveness and ease of use of our work, we chose Vim as the base model, which to our knowledge is the first pure-SSM-based visual model. Based on the open-source code provided by the authors, our method, called Vim-F, can be easily implemented by adding a few lines of code. Secondly, we improve the patch embedding to capture more local dependencies. Many successful lightweight transformer models, such as EfficientFormer \cite{ref11} and MobileViT \cite{ref12}, start their networks with a convolutional stem as patch embedding. This is attributed to the inherent spatial inductive bias of CNN, which is beneficial for capturing local information. In addition, CNNs are less sensitive to data augmentation, and the training process of a hybrid architecture is more stable. Inspired by these works, we design a patch embedding for Vim-F. Specifically, we use multiple convolutional layers for downsampling and then flatten the feature maps. Finally, we argue that the Mamba model is essentially still a recurrent neural network (RNN). Therefore, Vim lacks the rationale for using positional embeddings, and thus our Vim-F does not employ positional embeddings. Additionally, we reduce the model's sensitivity to the order of input sequences by alternately stacking linear attention blocks and Vim-F blocks, while maintaining approximately linear complexity. We refer to this hybrid architecture as Vim-F(H). Finally, we explore the significance and feasibility of the convolution-free visual Mamba encoder and propose a proof-of-concept network, Vim-F(CF). Through extensive experiments on image classification, object detection, instance segmentation, and semantic segmentation tasks, the simple Vim variants, Vim-F(H) and Vim-F(CF), achieve better performance. 
\section{Related Works}
\textbf{State Space Models.}
In mathematics,  State Space Models (SSMs) are usually expressed as linear ordinary differential equations (ODEs). These models transform input $D$-dimensional sequence $x(t) \in \mathbb{R}^{L \times D}$   into output sequence $y(t) \in \mathbb{R}^{L \times D}$ by utilizing a learnable latent state $h(t) \in \mathbb{R}^{L \times N}$ that is not directly observable. The mapping process could be denoted as:
\begin{align}\label{eq1}
	\begin{split}
		h'(t) &= \bm{A}h(t) + \bm{B}x(t),\\
		y(t) &= \bm{C}h(t), 
	\end{split}
\end{align}
where $\bm{A}\in\mathbb{R}^{N\times N}$, $\bm{B}\in\mathbb{R}^{D\times N}$ and $\bm{C}\in\mathbb{R}^{N \times D}$.\par
\textbf{Discretization.}
The primary objective of discretization is to convert ODEs into discrete functions, allowing the model to align with the sampling frequency of the input signal for more efficient computation. Following the work \cite{ref28}, the continuous parameters ($\bm{A}$, $\bm{B}$) can be discretized by zero-order hold rule with a given sample timescale $\Delta \in \mathbb{R}^{N}$:

\begin{equation} \label{eq2}
	\begin{aligned}
		&\bar{\bm{A}} = e^{\Delta \bm{A}}, \\
		&\bar{\bm{B}} = (\Delta\bm{A})^{-1}(e^{\Delta \bm{A}} - \bm{I})\cdot\Delta\bm{B}, \\
		&\bar{\bm{C}} = \bm{C}, \\
		&h(t) = \bar{\bm{A}}h(t-1) + \bar{\bm{B}}x(t), \\
		&y(t) = \bar{\bm{C}}h(t), 
	\end{aligned}
\end{equation}
where $\bm{\bar{A}}\in\mathbb{R}^{N\times N}$, $\bm{\bar{B}} \in \mathbb{R}^{D\times N}$ and $\bm{\bar{C}}\in\mathbb{R}^{N \times D}$. 

To improve computational efficiency, the iterative process described in equation \ref{eq2} can be accelerated through parallel computing using global convolutional operations:
\begin{align}
	\begin{split}
		\bm{y} &= \bm{x} \circledast \bm{\overline{K}} \\
		\text{with} \quad \bm{\overline{K}} &= (\bm{C}\bm{\overline{B}},\bm{C}\overline{\bm{A}\bm{B}}, ..., \bm{C}\bm{\overline{A}}^{L-1}\bm{\overline{B}}),
	\end{split}
\end{align}
where $\circledast$ denotes convolution operation, and $\bm{\overline{K}} \in \mathbb{R}^{L}$ is the SSM kernel.\par

\textbf{Selective State Space Models (S6).}
Mamba \cite{ref5} improves the performance of SSM by introducing Selective State Space Models (S6), allowing the continuous parameters to vary with the input enhances selective information processing across sequences, which extend the discretization process by selection mechanism:
\begin{equation}
	\begin{aligned}\label{eq4}
		\bar{\bm{B}} &= s_{\bm{B}}(x), \\
		\bar{\bm{C}} &= s_{\bm{C}}(x), \\
		\Delta &= softplus\big(\text{Parameter} + s_{\bm{A}}(x)\big), 
	\end{aligned}
\end{equation}
where $ s_{\bm{B}}(x) $ and $ s_{\bm{C}}(x) $ are linear functions that project input $ x $ into an $N$-dimensional space, while $ s_{\bm{A}}(x) $ broadens a $D$-dimensional linear projection to the necessary dimensions. Parameter represents a trainable parameter matrix. $softplus$ is a mathematical operation defined as: $softplus(x) = log(1 + e^x)$. In this paper, we use the term state space model (SSM) to refer to variants of SSMs, including S6.\par

\begin{figure*}[t]
	\centering
	\includegraphics[width=\textwidth]{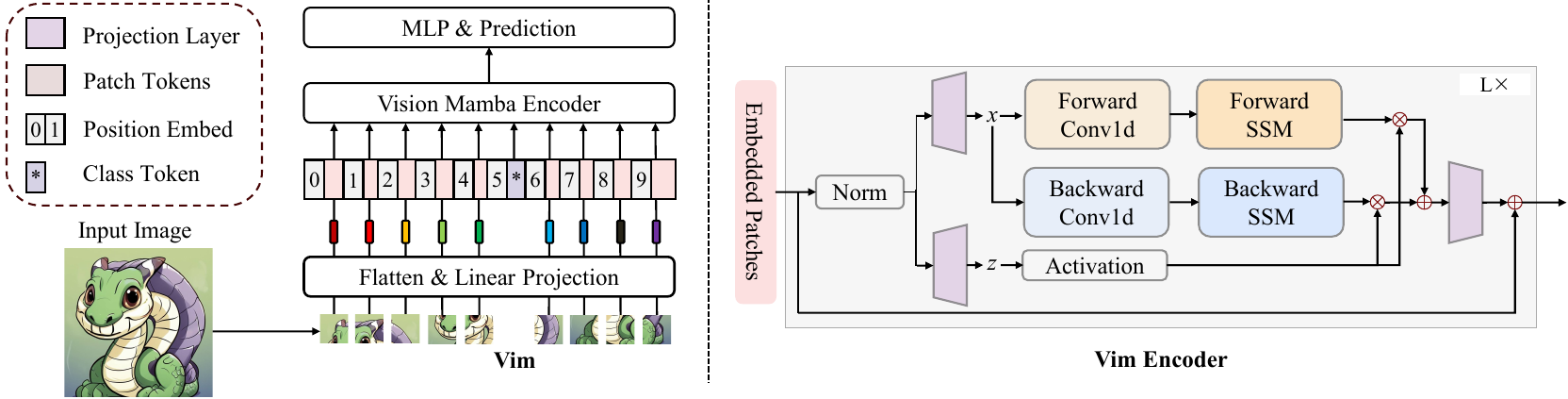}
	\caption{The overview of the Vim model.}
	\label{Fig1}
\end{figure*} 

\textbf{Overview of Vim.}
As mentioned earlier, the effectiveness of our work is demonstrated through the improvement of Vim. Therefore, we provide a brief introduction to the overall architecture of Vim. As shown in Fig.~\ref{Fig1}, Vim first reshapes the 2D image $\mathbf{X} \in \mathbb{R}^{h\times w\times c}$ into a sequence of flattened 2D patches  $\mathbf{X}_p \in \mathbb{R}^{n\times (p^2\cdot c)}$, where $c$ is the number of channels, $(h,w)$ is the resolution of the original image, and $(p,p)$ is the resolution of each image patch. The effective sequence length for Vim is therefore $n = hw/p^2$. Then, Vim linearly projects $\mathbf{X}_{p}$ into vectors of size $d$ and adds them to the position embeddings $E_{pos}\in \mathbb{R}^{(n+1)\times d}$, as follows:

\begin{equation}
	\begin{aligned}
		\label{eq5}
		\mathbf{T}_0 &= [\mathbf{X}_\mathtt{cls};\mathbf{X}_p^1\mathbf{W};\mathbf{X}_p^2\mathbf{W};\cdots;\mathbf{X}_p^i\mathbf{W};\cdots;\mathbf{X}_p^{d}\mathbf{W}] + \mathbf{E}_{pos}, \\
	\end{aligned}
\end{equation}
where $\mathbf{X}_p^{i}$ is the $i$-th patch of $\mathbf{X}$, $\mathbf{W} \in \mathbb{R}^{(p^2 \cdot c) \times d}$ is the learnable projection matrix. Vim uses class token to represent the whole patch sequence, which is denoted as $\mathbf{X}_\mathtt{cls}$. Vim then sends the token sequence ($\mathbf{T}_{l-1}$) to the $l$-th layer of the Vim encoder, and gets the output $\mathbf{T}_{l}$. Finally, Vim normalizes the output class token $\mathbf{T}_{L}^0$ and feeds it to the multi-layer perceptron (MLP) head to get the final prediction $\hat{p}$, as follows:

\begin{equation}
	\begin{aligned}
		\mathbf{T}_l &= \mathbf{Vim}(\mathbf{T}_{l-1}) + \mathbf{T}_{l-1}, \\
		\mathbf{\it f} &= \mathbf{Norm}(\mathbf{T}_{L}^0), \\
		\hat{p} &= \mathbf{MLP}(\mathbf{\it f}), \\
	\end{aligned}
\end{equation}

where $\mathbf{Vim}$ is the proposed vision mamba block, $L$ is the number of blocks, and $\mathbf{Norm}$ is the normalization layer. The main difference between the Vim block and the classic Mamba block \cite{ref5} resides in that Vim introduces a bidirectional scanning strategy, employing two independent branches to process the forward and backward sequences respectively. Each branch comprises a convolutional layer and an S6 module.

\section{Rethinking the Scanning Strategy of ViMs}
The non-causal nature of two-dimensional spatial patterns in images is inherently inconsistent with the causal processing framework of SSM. Therefore, it is meaningful to explore methods for serializing images. However, most existing works on scanning strategies are typically based on a rather idealized assumption, that is, introducing more or more complex one-dimensional sequence scanning strategies may help enhance Mamba's ability to understand local two-dimensional correlations in images. However, most existing works on scanning strategies are usually based on a rather idealized assumption. Scanning the same sequence multiple times in different orders can expand Mamba's receptive field in different directions, thereby enhancing Mamba's ability to understand local 2D correlations in images. Theoretically, regardless of the order, reducing the distance between adjacent tokens in one direction in two-dimensional space inevitably increases the distance between adjacent tokens in another direction, thus these strategies are essentially equivalent. Additionally, in the absence of global information serving as a reference, these sorting methods are essentially disconnected, making it inherently difficult for Mamba to gain additional spatial relationship information by scanning multiple sequences.\par 
In practice, most works involving scanning strategy research, including Vim \cite{ref1}, LocalMamba \cite{ref3}, EfficientVMamba \cite{ref4}, and PlainMamba \cite{ref36}, lack clear comparative experiments to verify the necessity and effectiveness of their scanning strategies. VMamba \cite{ref2} conducted experiments on different scanning strategies using ImageNet-1K, and the results indicate that most strategies show no significant performance differences. Zhu et al. \cite{ref37} conducted a comprehensive experimental investigation on the impact of mainstream scanning directions and their combinations on semantic segmentation of remotely sensed images. Through extensive experiments on the LoveDA, ISPRS Potsdam, and ISPRS Vaihingen datasets, they demonstrate that no single scanning strategy outperforms others, regardless of their complexity or the number of scanning directions involved. In summary, an effective scanning strategy needs to maintain a global receptive field throughout the scanning process and demonstrate its effectiveness through analysis and comparative experiments.

\section{Methodology}
In this section, we first elaborate on the advantages of frequency-domain scanning and the scanning strategy we proposed. Subsequently, we present the specific details of patch embedding in Vim-F. We then discuss the composition of the Vim-F and Vim-F(H) blocks. Finally, we explore the significance and feasibility of the convolution-free visual Mamba encoder and propose a proof-of-concept network, Vim-F(CF).
\subsection{Scanning in the Frequency Domain}
As analyzed in Section 3, maintaining a global receptive field during the scanning process is crucial for enhancing Mamba's ability to understand local 2D correlations in images through improved scanning strategies. We believe that introducing frequency-domain scanning is a simple and straightforward solution. Specifically, we first perform a Fourier transform on the feature map to obtain:
\begin{equation}\label{eq7}  
	F(u, v) = \sum_{x=0}^{H-1} \sum_{y=0}^{W-1} f(x, y) e^{-j2\pi\left(\frac{ux}{H} + \frac{vy}{W}\right)} ,
\end{equation}
where $j$ represents the imaginary unit, $H$ and $W$ represent the height and width of the feature map, respectively. $f(x,y) \in \mathbb{R}^{H\times W}$ represents the corresponding value at the coordinate $(x,y)$ in the feature map, and $F(u,v)\in \mathbb{R}^{H\times W}$ represents the corresponding value at the coordinate $(u,v)$ in the frequency domain. From Eq. (\ref{eq7}), it can be seen that each frequency-domain coefficient $F(u,v)$ is related to the values of all pixels in the feature map. $|F(u,v)|$ represents the modulus of $F(u,v)$, which is also known as the amplitude spectrum. Obviously, scanning on the amplitude spectrum can ensure a global receptive field throughout the scanning process. \par
\begin{figure*}[t]
	\centering
	\includegraphics[width=\textwidth]{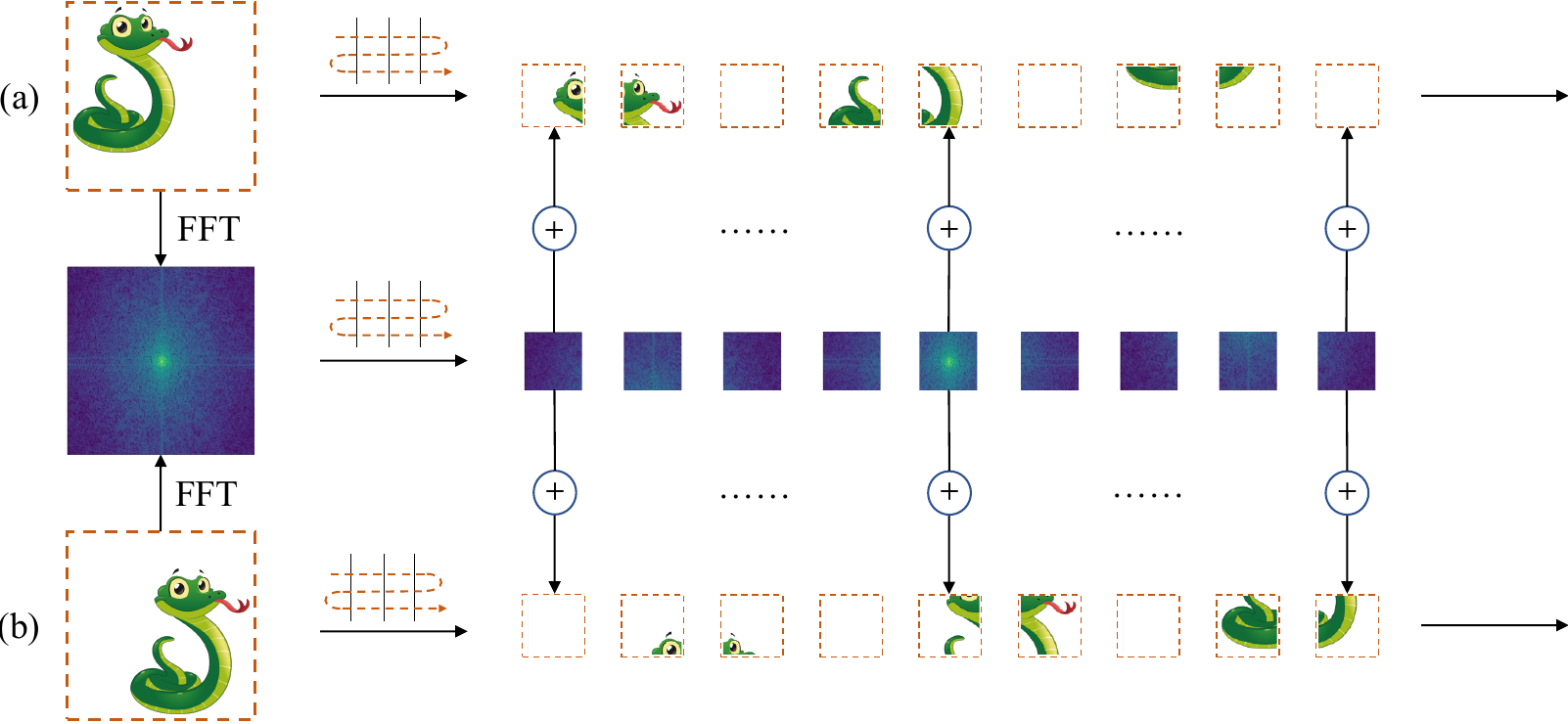}
	\caption{Illustrating the scanning strategy of Vim-F through an example of forward scanning in both the frequency and spatial domains. Due to the translation invariance of the Fourier transform, the amplitude spectra of figures (a) and (b) are identical.}
	\label{Fig2}
\end{figure*}

In addition, the Fourier transform has translation invariance. Specifically, if a function is translated in the spatial domain, its amplitude spectrum is the same as that of the original function. A brief proof is provided below:
According to Eq. (\ref{eq7}), assuming the translated feature map is $f(x-a,y-b)$, where $a$ and $b$ are the translation amounts. The Fourier transform of the translated feature map is:
\begin{equation}\label{eq8}  
	F'(u, v) = \sum_{x=0}^{H-1} \sum_{y=0}^{W-1} f(x-a, y-b) e^{-j2\pi\left(\frac{ux}{H} + \frac{vy}{W}\right)} .
\end{equation}
Let $x'=x-a$, $y'=y-b$, then $x=x'+a$, $y=y'+b$. Substituting into Eq. (\ref{eq8}), we get: 
\begin{equation}\label{eq9}
	\begin{aligned}
	     F'(u, v) & = \sum_{x'} \sum_{y'} f(x', y') e^{-j2\pi\left(\frac{u(x'+a)}{H} + \frac{v(y'+b)}{W}\right)} \\     
	              & = e^{-j2\pi\left(\frac{ua}{H} + \frac{vb}{W}\right)}\sum_{x'} \sum_{y'} f(x', y') e^{-j2\pi\left(\frac{ux'}{H} + \frac{vy'}{W}\right)} \\     
	              & = e^{-j2\pi\left(\frac{ua}{H} + \frac{vb}{W}\right)}F(u,v) .\\  
	\end{aligned}
\end{equation}
We know that the modulus of the phase factor $e^{-j2\pi\left(\frac{ua}{H} + \frac{vb}{W}\right)}$ is 1, so the amplitude spectrum of $F'(u, v)$ is:
\begin{equation}\label{eq10}  
|F'(u, v)| = |e^{-j2\pi\left(\frac{ua}{H} + \frac{vb}{W}\right)}F(u,v)| = |F(u,v)|.
\end{equation}
This property makes the amplitude spectrum an ideal ``reference object" during scanning, so simultaneous scanning in both the frequency and spatial domains is better than scanning in either domain alone. As shown in Fig.~\ref{Fig2}, adding the amplitude spectrum to the original feature map introduces global information while reducing inductive biases from one-dimensional serialization, thereby enhancing Mamba's ability to understand semantic objects.

\subsection{Patch Embedding}
In Vim and ViTs, the patching strategy typically involves using large convolutional kernels (such as kernel sizes of 14 or 16) with non-overlapping convolutions. However, VMamba and Swin Transformer utilize non-overlapping convolutions with a kernel size of 4, indicating that current ViMs haven't undergone much targeted design in terms of patching. We believe that non-overlapping convolutions have fewer inductive biases, and the self-attention mechanism of the Transformer architecture excels at modeling relationships between patches. Therefore, adopting non-overlapping convolutions can benefit the performance of Transformers. On the contrary, the additional correlation between patches introduced by overlapping convolutions can be advantageous for the scanning mechanism of ViMs.\par
Based on the above analysis, we can try to explain why Vim uses the position embedding. The bidirectional horizontal scanning strategy employed by Vim makes adjacent tokens in horizontal and vertical directions far apart in the sequence. In addition, the lack of data correlation between tokens makes it difficult for the model to understand the spatial relationship in the vertical direction. The introduction of the trainable position embedding can explicitly provide the model with the absolute position of each token in the plane space, thus alleviating the above problems. In contrast, it is simpler and more flexible to remove the position embedding and establish spatial correlation through overlapping convolutions, enabling the model to learn spatial positional relationships.\par
Specifically, we first utilize a $7\times7$ convolutional layer with a stride of 4 to model the local correlation of the input image. Then, we downsample the feature map twice through two downsampling blocks. Each downsampling block consists of a $2\times2$ convolutional layer with a stride of 2 and a $1\times1$ convolutional layer. The details of the adaptation with the two base models \cite{ref1} (Vim-Ti and Vim-S) are shown in Table \ref{tab1}.
\begin{table}[htbp]
	\centering
	\renewcommand{\arraystretch}{1.1}
	\caption{Convolutional stem for patch embedding.}
	\begin{tabular}{cccc}
		& \begin{tabular}[c]{@{}c@{}} Output size\end{tabular} &  Vim-Ti & Vim-S   \\
		\hline
		\multirow{2}{*}{Block1} & \multirow{2}{*}{\begin{tabular}[c]{@{}c@{}} 56$\times$56 \end{tabular}} & 
		7$\times$7, 48, stride 4  
		& 7$\times$7, 48, stride 4  \\ 
		
		& & padding 3 & padding 3 \\
		
		\hline
		\multirow{4}{*}{Block2} & 
		\multirow{4}{*}{\begin{tabular}[c]{@{}c@{}} 28$\times$28 \end{tabular}} & 
		\multirow{4}{*}{$\begin{bmatrix}\text{2$\times$2, 96, stride 2}\\\text{1$\times$1, 96}\end{bmatrix}$}  & 
		\multirow{4}{*}{$\begin{bmatrix}\text{2$\times$2, 192, stride 2}\\\text{1$\times$1, 192}\end{bmatrix}$}  \\
		& & &  \\
		& & &  \\
		& & &  \\

		\hline
		\multirow{4}{*}{Block3} & 
		\multirow{4}{*}{\begin{tabular}[c]{@{}c@{}} 14$\times$14 \end{tabular}} & 
		\multirow{4}{*}{$\begin{bmatrix}\text{2$\times$2, 192, stride 2}\\\text{1$\times$1, 192}\end{bmatrix}$}&
		\multirow{4}{*}{$\begin{bmatrix}\text{2$\times$2, 384, stride 2}\\\text{1$\times$1, 384}\end{bmatrix}$} \\
		& & &  \\
		& & &  \\
		& & &  \\
		\hline
		
	\end{tabular}
	\normalsize
	\label{tab1}
\end{table}\par
The patch embedding in Vim can be implemented through a $16\times 16$ convolutional layer with a stride of 16. The convolutional stem for patch embedding in Vim-F comprises more convolutional layers, but the additional computational cost is negligible due to the use of smaller convolutional kernels. Considering Vim-Ti as the baseline model, the computational load measured in GFLOPs only increases by 0.035. 

\subsection{Vim-F Block}
\begin{figure}[h]
	\centering
	\includegraphics[width=0.85\columnwidth]{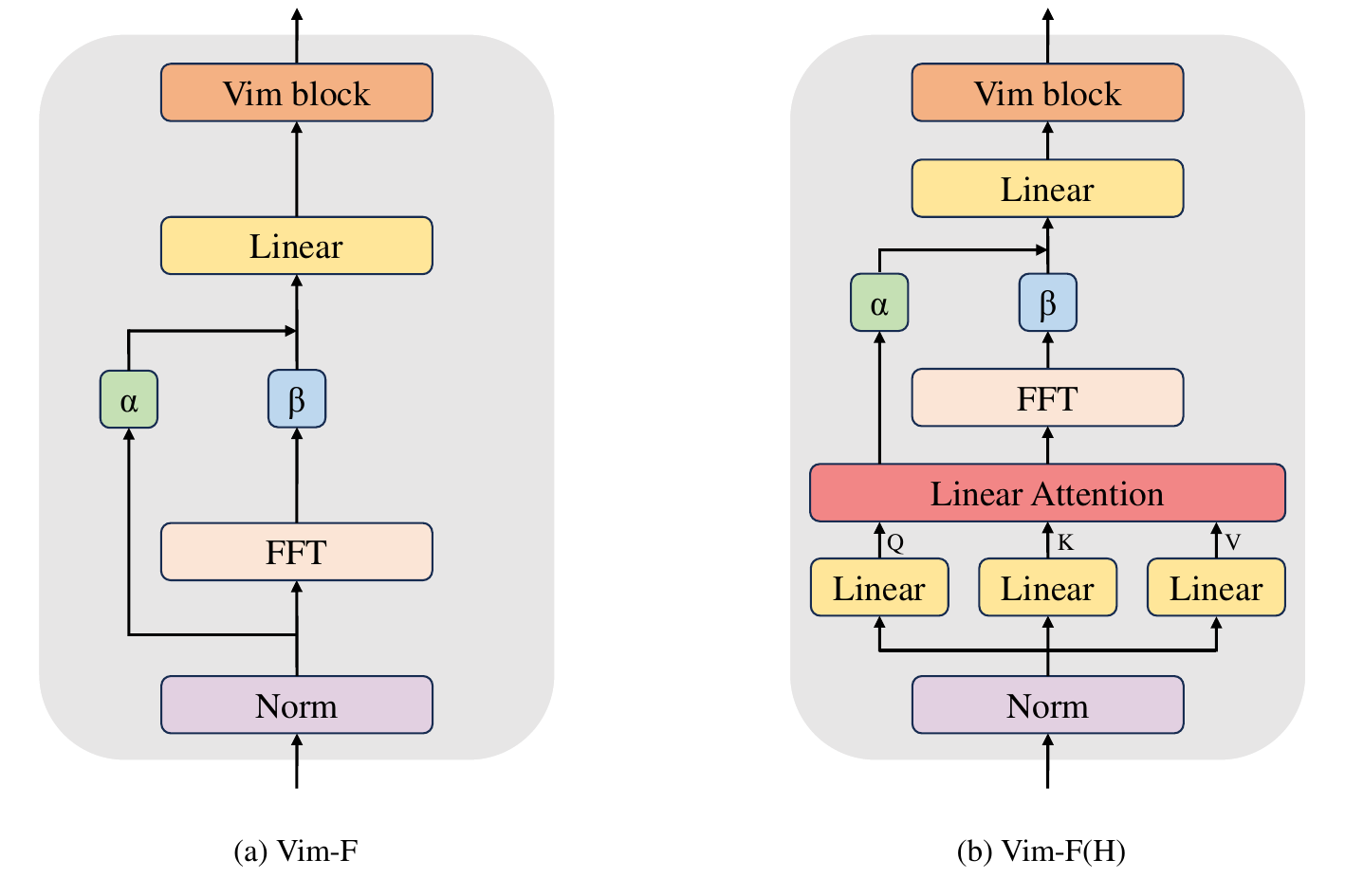}
	\caption{The architecture of the Vim-F block and the Vim-F(H) block.}
	\label{Fig6}
\end{figure}

As mentioned earlier, the Vim-F block is an improvement over the Vim \cite{ref1} block, as shown in Fig.~\ref{Fig6}(a). Specifically, a 2D FFT is performed on the input to obtain the spectrogram. Then, the spectrogram is added to the original input, with the intensity of the frequency and spatial domain information being adjusted by two trainable parameters, $\alpha$ and $\beta$, respectively. After the linear layer mixes the hidden dimensions of the feature map, the cross-domain information is sent to the Vim block for processing. Vim-F requires a 2D discrete Fourier transform (DFT) to be performed on the feature map before scanning, thus introducing additional computation. The Cooley-Tukey FFT algorithm decomposes a DFT of a sequence with length $N$ into two DFTs of sub-sequences with length $N/2$, significantly reducing the computational complexity. During the execution of the algorithm, the data is divided into two parts based on whether they are odd or even, and the DFTs of each part are computed separately before merging the results. For a 2D-DFT with dimensions $(L, D)$, its complexity is $\Omega(L\cdot D \cdot logL\cdot logD)$. The complexity of the linear layer is $\Omega(L\cdot D^2)$. Therefore, compared to the Vim block, the additional overhead of the Vim-F block is approximately linearly related to the sequence length $L$. Considering that Vim \cite{ref1} flattens the input into a sequence of length 196, the hidden state dimensions $D$ of the two variants Vim-Ti and Vim-S are 192 and 384 respectively, with a state expansion ratio of 2. These additional computational overheads are even significantly smaller than any single linear projection in Eq. (\ref{eq4}).
\subsection{Vim-F(H) Block}
Observing Eq. (\ref{eq2}), it is evident that since all elements of matrix $A$ are strictly confined within the range of 0 to 1 \cite{ref5}, the early information stored in the hidden state $h$ will continuously decay. This leads to Mamba being sensitive to the order of the input sequence. We propose that by weighting the input sequence with attention mechanisms before processing with Mamba, this strong local bias can be mitigated, thereby enhancing Mamba's capability to handle long sequences. We integrate the Vim-F block with the linear attention \cite{ref38} to maintain approximately linear complexity, which we term Vim-F(H), and its structure is depicted in Fig.~\ref{Fig6}(b).
\begin{figure}[h]
	\centering
	\includegraphics[width=\columnwidth]{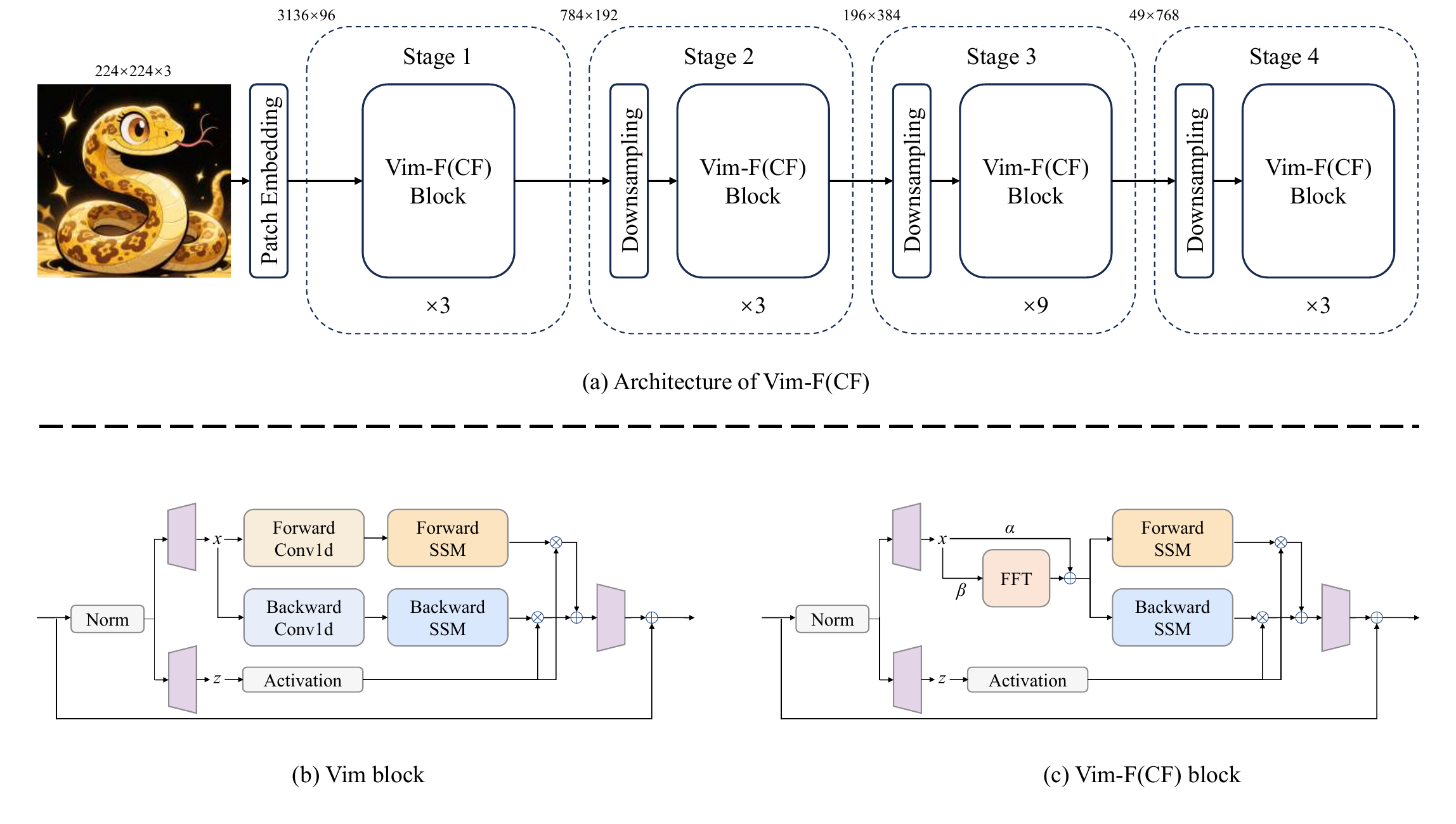}
	\caption{Illustration of (a) Overall architecture of Vim-F(CF), (b) Vim block structure for reference, and (c) Proposed Vim-F(CF) block structure.}
	\label{Fig7}
\end{figure}
\subsection{Vim-F(CF): Is the convolution necessary in the ViM encoder?}
We know that Mamba \cite{ref5} was originally designed for NLP, with its core module S6 being good at capturing long-range dependencies. Convolutions were introduced to compensate for the model's limitations in short-range context modeling. However, the sequence lengths processed by ViMs are usually short. For example, in Vim \cite{ref1} and PlainMamba \cite{ref36}, images are flattened into sequences with a length of 196. Thus, from this perspective, retaining convolutions in the ViM encoder is unnecessary. Due to the influence of traditional visual model design concepts and experimental results, the majority of existing ViM encoders retain convolutional layers. VMamba \cite{ref2} points out that depthwise convolutions play an important role in forming a global effective receptive field (ERF). We believe that removing the convolutional layer and adopting frequency domain scanning can make the modeling approach of ViM closer to that of Transformer, thereby enhancing the scalability and transferability of ViM.\par
We use convolution-free Vim-F blocks to build the proof-of-concept network Vim-F(CF), as shown in Fig.~\ref{Fig7}. Vim-F(CF) follows a classic 4-stage structure to enable SSM to model visual features across different scales. Unlike common CNNs, Transformers, ViMs, and their hybrid models, Vim-F(CF) does not incorporate attention mechanisms or feed forward network (FFN) modules.
\section{Experiments}
This section presents our experimental results, starting with the ImageNet classification task and then transferring the trained model to various downstream tasks, including object detection, instance segmentation, and semantic segmentation. The specific architectures of Vim-F and Vim-F(H) are as follows: We first replace the patch embedding module, and then only the first 6 Vim blocks (25\% of the total number of blocks) are substituted with either Vim-F or Vim-F(H) blocks to balance performance and efficiency. The number of linear attention heads in Vim-F(H) is set to 24.
\definecolor{Grey}{RGB}{240,240,240}
\begin{table}[h]
	\centering
	\renewcommand{\arraystretch}{1.1}
	\caption{Comparison of different backbones on ImageNet-1K classification.}
	\begin{tabular}{lccc}
		\hline
		Method & Params (M)\; & FLOPs (G)\; & Top-1 (\%)\;\\
		\hline
		PVTv2-B0 \cite{ref31} & 3 & 0.6 & 70.5\\
		DeiT-Ti \cite{ref32} &  6 & 1.3 & 74.4\\
		MobileViT-XS \cite{ref12}  & 2 & 1.0 & 74.8\\
		LVT \cite{ref34} & 6 & 0.9 & 74.8\\
		Vim-Ti \cite{ref1} & 7 & 1.5 & 76.1\\
		LocalVim-T \cite{ref3} & 8 & 1.5 & 76.5\\
		Vim-Ti-F \textbf{ (ours)}  & 7 & 1.5 & 76.7\\
		PlainMamba-L1 \cite{ref36} & 7 & 3.0 & 77.9\\
		RegNetY-1.6G \cite{ref20}  & 11 & 1.6 & 78.0\\
		Vim-Ti-F(H) \textbf{ (ours)}  & 9 & 2.1 & 78.2\\
		MobileViT-S \cite{ref12} & 6 & 2.0 & 78.4\\
		EfficientVMamba-S \cite{ref4}  & 11 & 1.3 & 78.7 \\
		DeiT-S \cite{ref32}  & 22 & 4.6 & 79.8\\
		RegNetY-4G \cite{ref20}  & 21 & 4.0 & 80.0\\
		Vim-S \cite{ref1} & 26 & 5.1 & 80.5\\
		Vim-F(CF) \textbf{ (ours)}  & 22 & 5.2 & 80.7\\
		Vim-S-F \textbf{ (ours)}  & 28 & 5.5 & 80.9\\
		LocalVim-S\cite{ref3}  & 28 & 4.8 & 81.0\\
		Swin-T \cite{ref22} & 29 & 4.5 & 81.3\\
		PlainMamba-L2 \cite{ref36} & 25 & 8.1 & 81.6\\
		Vim-S-F(H) \textbf{ (ours)}  & 37 & 6.8 & 81.6\\
		EfficientVMamba-B \cite{ref4}  & 33 & 4.0 & 81.8 \\
		VMamba-T \cite{ref2}  & 30 & 4.9 & 82.6\\
		\hline
	\end{tabular}
	\label{tab2}
\end{table}
\subsection{Image Classification}
\subsubsection{Settings. }We train the models on ImageNet-1K and evaluate the performance on ImageNet-1K validation set. For fair comparisons, our training settings mainly follow Vim. Due to hardware limitations, our experiments are performed on 3 A6000 GPUs. Therefore, we adjust the total batch size and the initial learning rate to 384 and $3.75 \times 10^{-4} $ respectively according to the linear scaling rule\cite{ref30}. \par
\subsubsection{Results. }We selected advanced CNNs, ViTs, and ViMs with comparable parameters and computational costs in recent years to compare with our method, and the results are shown in Table \ref{tab2}. Vim-F outperforms the original Vim in terms of performance, and Vim-F(H) further enhances Vim-F's performance by introducing linear attention. Although not meticulously designed, the proof-of-concept network Vim-F(CF) matches Vim-S in performance and computational complexity and has a parameter count advantage, indicating that convolution-free ViM encoders and associated model architectures are also a promising research direction. It is worth noting that SOTA ViMs such as VMamba \cite{ref2}, PlainMamba \cite{ref36}, and EfficientVMamba \cite{ref4} all adopt more complex CNN-Mamba hybrid encoders, using depthwise separable convolutions to extract features in a 2D space. In addition, their network architectures also draw on advanced networks like ConvNeXt \cite{ref39}, MetaFormer \cite{ref40}, MobileViT \cite{ref12}. Since these works have not conducted ablation experiments on the necessity of the SSM block, it is reasonable to suspect that the exceptional performance of these ViMs may mainly be attributed to the introduction of depthwise separable convolutions and advanced network architectures.
Unlike these works, Vim-F and Vim-F(H) do not use depthwise separable convolutions and largely follow the design philosophy of Vim's network architecture. LocalVim\cite{ref3} is one of the representative works that improve the scanning strategy for Vim. LocalVim has designed 8 scanning strategies and employs neural architecture search to determine the scanning strategy used by each of the 4 scanning branches in every LocalVim block. Additionally, it utilizes a complex spatial and channel attention module (SCAttn) to weight the channels and tokens in each branch feature. Moreover, the total number of blocks in LocalVim has not followed Vim's 24 but has been reduced to 20. In comparison, our works achieve a better balance between performance, efficiency, and model complexity. \par
\begin{table}[htbp]
	\centering
	\renewcommand{\arraystretch}{1.1}
	\caption{Ablation study on the proposed patch embedding module and Vim-F block proportion.}
		\setlength{\tabcolsep}{7pt} 
	\begin{tabular}{lccc}
		\hline
		Method  & Params (M) & FLOPs (G)& Top-1 (\%)\\
		\hline
		Vim-Ti \cite{ref1}  & 7 & 1.5 & 76.1\\	
		\;- Position Embedding & 7 & 1.5 & 74.8\\	
		\hline		
		Vim-Ti-F-V0   & 7 & 1.5 & 76.3\\		
		Vim-Ti-F-V1   & 7 & 1.5 & 76.7\\
		Vim-Ti-F-V2   & 8 & 1.7 & 76.8\\
		Vim-Ti-F-V3   & 9 & 1.9 & 77.0\\	
		\hline
	\end{tabular}
	\label{tab4}
\end{table}
\subsubsection{Ablation Study on ImageNet-1K. }Here, taking Vim-Ti-F as an example, we conduct an ablation study on the effectiveness of our proposed patch embedding module and the impact of the Vim-F block proportion. We configured four variants of Vim-Ti-F, namely V0 to V3, with Vim-F block proportions set at 0\%, 25\%, 50\%, and 100\%, respectively. As shown in Table \ref{tab4}, removing position embedding caused a 1.3\% performance degradation of Vim-Ti. Vim-Ti-F-V0 showed slightly better performance than Vim-Ti, indicating that using a convolutional stem with overlapping convolutions as patch embedding is a more effective approach. Additionally, substituting more Vim blocks with Vim-F blocks can lead to steady performance improvement. However, due to the overall representation limitations of the model, it is unnecessary to use Vim-F blocks exclusively.
\begin{table*}[h]
	\caption{Object detection and instance segmentation results on COCO.} 
	\label{tab3}	 
	\centering  
	\renewcommand{\arraystretch}{1.1}  
	\begin{tabular}{lcccccccc} 
		\hline
		Backbone & Params & FLOPs & AP$^\mathrm{b}$ & AP${}^\mathrm{b}_{50}$ & AP${}^\mathrm{b}_{75}$ & AP${}^\mathrm{m}$ & AP${}^\mathrm{m}_{50}$ & AP${}^\mathrm{m}_{75}$\\
		\hline
		ResNet-18 \cite{ref15} \ & 31M & 207G & 34.0 & 54.0 & 36.7 & 31.2 & 51.0 & 32.7\\
		Vim-Ti \cite{ref1}     & 27M & 189G & 36.6 & 59.4 & 39.2 & 34.9 & 56.7 & 37.3\\
		Vim-Ti-F   \textbf{(ours)} & 28M & 189G & 37.7 & 60.8 & 40.9 & 35.7 & 57.8 & 38.0\\
		ResNet-50 \cite{ref15} & 44M & 260G & 38.0 & 58.8 & 41.4 & 34.7 & 55.7 & 37.2\\
		Vim-Ti-F(H)   \textbf{(ours)} & 28M & 189G & 39.3 & 61.8 & 42.6 & 36.7 & 58.9 & 39.2\\
		EfficientVMamba-S \cite{ref4} & 31M & 197G & 39.5 & 60.8 & 43.1 & 35.8 & 57.4 & 38.3\\ 
		ResNet-101 \cite{ref15}        & 63M & 336G & 40.0 & 60.5 & 44.0 & 36.1 & 57.5 & 38.6\\
		Vim-S \cite{ref1}            & 44M & 272G & 40.9 & 63.9 & 45.1 & 37.5 & 59.4 & 40.2\\
		Vim-S-F   \textbf{(ours)}    & 47M & 273G & 41.6 & 62.7 & 45.6 & 37.9 & 60.8 & 40.7\\		
		Vim-F(CF)   \textbf{(ours)} & 42M & 272G & 42.0 & 62.5 & 45.9 & 38.4 & 60.9 & 41.3\\
		Swin-T \cite{ref22}           & 48M & 267G & 42.7 & 65.2 & 46.8 & 39.2 & 62.1 & 42.1\\	 
		Vim-S-F(H)   \textbf{(ours)}  & 57M & 275G & 42.9 & 65.4 & 47.1 & 39.3 & 62.2 & 42.2\\
		\hline
	\end{tabular}  
\end{table*}
\subsection{Object Detection and Instance Segmentation}
\subsubsection{Settings. } We used Mask-RCNN as the detector to evaluate the performance of the proposed methods for object detection and instance segmentation on the MSCOCO 2017 dataset. Since the backbones of Vim-F and Vim-F(H) are non-hierarchical, following ViTDet\cite{ref35}, we only used the last feature map from the backbone and generated multi-scale feature maps through a set of convolutions or deconvolutions to adapt to the detector. All remaining settings were consistent with Swin\cite{ref22}. Specifically, we employed the AdamW optimizer and fine-tune the pre-trained classification models (on ImageNet-1K) for both 12 epochs (1$\times$ schedule). The learning rate is initialized to $1\times 10^{-4}$ and is reduced by a factor of $10\times$ at the 9th and 11th epochs.\par
\subsubsection{Results. }We summarize the comparison results of Vim-F, Vim-F(H), and Vim-F(CF) with other backbones in Table \ref{tab3}. It can be seen that our Vim-F consistently outperforms Vim. Notably, Vim-S-F outperforms Vim-F(CF) in classification tasks, while Vim-F(CF) surpasses Vim-S-F in downstream dense prediction tasks. Similar to the results in classification tasks, Vim-F(H) exhibits better performance compared to both Vim-F and Vim-F(CF), and it is competitive with state-of-the-art ViT architectures.

\subsection{Semantic Segmentation}
\textbf{Settings. }Following Vim \cite{ref1}, we train UperNet \cite{ref43} with our models on ADE20K dataset. In training, we employ AdamW with a weight decay of 0.01, and a total batch size of 16 to optimize models. The employed training schedule uses an initial learning rate of $6 \times 10^{-5}$, linear learning rate decay, a linear warmup of 1500 iterations, and a total training of 160K iterations. \par
\begin{table}[h] 
	\renewcommand{\arraystretch}{1.1}  
	\centering  
	\caption{Results of semantic segmentation on ADE20K.} 
		\setlength{\tabcolsep}{12pt} 
	\begin{tabular}{lcc} 
		\hline
		Backbone & Params & mIoU\\
		\hline
		Vim-Ti \cite{ref1}            & 34M & 40.1 \\
		Vim-Ti-F   \textbf{(ours)}    & 34M & 40.4 \\
		ResNet-50 \cite{ref15}        & 67M & 40.7 \\
		Vim-S \cite{ref1}             & 57M & 43.9 \\
		Vim-S-F   \textbf{(ours)}     & 58M & 44.2 \\
		Vim-F(CF)   \textbf{(ours)}   & 55M & 44.4 \\
		Swin-T \cite{ref22}           & 60M & 44.5 \\ 
		Vim-S-F(H)   \textbf{(ours)}  & 47M & 44.8 \\
		\hline
	\end{tabular}  
	\label{tab7}	
\end{table}
\textbf{Results. }The results are presented in Table \ref{tab7}. 
Similar to the experimental results on MSCOCO, Vim-F and its variants demonstrate superior performance to Vim \cite{ref1}, ResNet \cite{ref15}, and Swin\cite{ref22} in the semantic segmentation task, which further confirms the effectiveness of our work.
\section{Conclusion}
One of the main challenges currently faced by ViMs is the destruction of local two-dimensional dependency relationships when spatial data is flattened into one-dimensional tokens due to the non-causal nature of visual data. Consequently, scanning one-dimensional tokens with specific rules can limit the model's ability to accurately interpret spatial relationships. We observe that scanning in the frequency domain, leveraging the translation invariance and global nature of Fourier transforms, can reduce the inductive bias of scanning strategies while ensuring that the model always maintains a good global receptive field. We adapt a simple approach of adding the spectrogram with the original feature map, allowing the model to scan simultaneously in both the frequency and spatial domains, thereby establishing a unified representation of visual information across both domains. Furthermore, we employ two trainable parameters to dynamically adjust the information intensity in the frequency and spatial domains, aiming to reduce semantic gaps. We also introduce additional local correlations during patch embedding by using overlapping convolutions. Vim-F(H) enhances the performance of Vim-F by introducing linear attention, which mitigates the sequence order sensitivity of the Mamba encoder. We constructed a proof-of-concept network, Vim-F(CF), and found that even after removing the convolutional layers from Vim-F, the model can still effectively model visual information. Extensive experiments show that our work effectively boosts Vim's performance. Vim-F, Vim-F(H), and Vim-F(CF) achieve results competitive with advanced ViM models.

%
%
%
\bibliographystyle{splncs04}
\bibliography{ref}
\end{document}